\documentclass[lettersize,journal]{IEEEtran}
\usepackage{amsmath,amsfonts}
\usepackage{algorithm}
\usepackage{array}
\usepackage[caption=false,font=normalsize,labelfont=sf,textfont=sf]{subfig}
\usepackage{textcomp}
\usepackage{stfloats}
\usepackage{url}
\usepackage{verbatim}
\usepackage{graphicx}
\usepackage{cite}

\usepackage{subcaption} 
\usepackage[table]{xcolor} 
\usepackage{nicefrac}       
\usepackage{microtype}      
\usepackage{marvosym}
\usepackage{subfig}
\usepackage{amssymb}
\usepackage{makecell}
\usepackage{xcolor}
\usepackage{algpseudocode}
\usepackage{multirow}
\usepackage{mathtools} 
\usepackage{booktabs} 
\usepackage{amsmath} 
\usepackage{amsthm} 
\usepackage{color} 
\usepackage{colortbl} 
\usepackage{bm} 
\usepackage{float} 
\definecolor{rblue}{rgb}{0,0.5,1}
\usepackage[bookmarks=false]{hyperref}
\usepackage[capitalise]{cleveref}
\hypersetup{colorlinks=true,linkcolor=red,citecolor=rblue}

\begin{document}

\title{Scene Graph-guided SegCaptioning Transformer with Fine-grained Alignment for Controllable Video Segmentation and Captioning}

\author{
Xu~Zhang\IEEEauthorrefmark{1},
Jin~Yuan\IEEEauthorrefmark{1},
BinHong~Yang,
Xuan Liu,
Qianjun Zhang\IEEEauthorrefmark{2},
Yuyi Wang\IEEEauthorrefmark{2},
Zhiyong~Li,
and Hanwang~Zhang
\thanks{This work was supported in part by the National Natural Science Foundation of China (No.62406263, No. 62272157, and No. U23A20341).}
\thanks{X. Zhang, B. Yang, J. Yuan, and X. Liu are with the College of Computer Science and Electronic Engineering, Hunan University, Changsha 410082, China.}
\thanks{Z. Li is with the School of Robotics and the National Engineering Research Center of Robot Visual Perception and Control Technology, Hunan University, Changsha 410082, China.}
\thanks{Q. Zhang is with the School of Computing and Artificial Intelligence, Southwest Jiaotong University, Chengdu 611756, P.R. China.}
\thanks{Y. Wang is with CRRC Zhuzhou Institute Company Ltd., Zhuzhou, Hunan 412001, China.}
\thanks{H. Zhang is with Nanyang Technological University, Singapore 639798.}
\thanks{\IEEEauthorrefmark{2}Corresponding authors: Qianjun Zhang and Yuyi Wang. (E-mail: zqjblue@foxmail.com, yuyiwang920@gmail.com.)}
\thanks{\IEEEauthorrefmark{1}Equal contribution.}
}

\maketitle

\begin{abstract}
Recent advancements in multimodal large models have significantly bridged the representation gap between diverse modalities, catalyzing the evolution of video multimodal interpretation, which enhances users' understanding of video content by generating correlated modalities. However, most existing video multimodal interpretation methods primarily concentrate on global comprehension with limited user interaction. To address this, we propose a novel task, Controllable Video Segmentation and Captioning (SegCaptioning), which empowers users to provide specific prompts, such as a bounding box around an object of interest, to simultaneously generate correlated masks and captions that precisely embody user intent. An innovative framework Scene Graph-guided Fine-grained SegCaptioning Transformer (SG-FSCFormer) is designed  that integrates a Prompt-guided Temporal Graph Former to effectively captures and represents user intent through an adaptive prompt adaptor, ensuring that the generated content well aligns with the user’s requirements. Furthermore, our model introduces a Fine-grained Mask-linguistic Decoder to collaboratively predict high-quality caption-mask pairs using a Multi-entity Contrastive loss, as well as provide fine-grained alignment between each mask and its corresponding caption tokens, thereby enhancing users' comprehension of videos.  
Comprehensive experiments conducted on two benchmark datasets demonstrate that SG-FSCFormer achieves remarkable performance, effectively capturing user intent and generating precise multimodal outputs tailored to user specifications.
Our code is available at~\url{https://github.com/XuZhang1211/SG-FSCFormer}. 

\end{abstract}



\section{Introduction}
\label{sec:intro}

\begin{figure}[t!]
	\centering
	\includegraphics[width=1\linewidth]{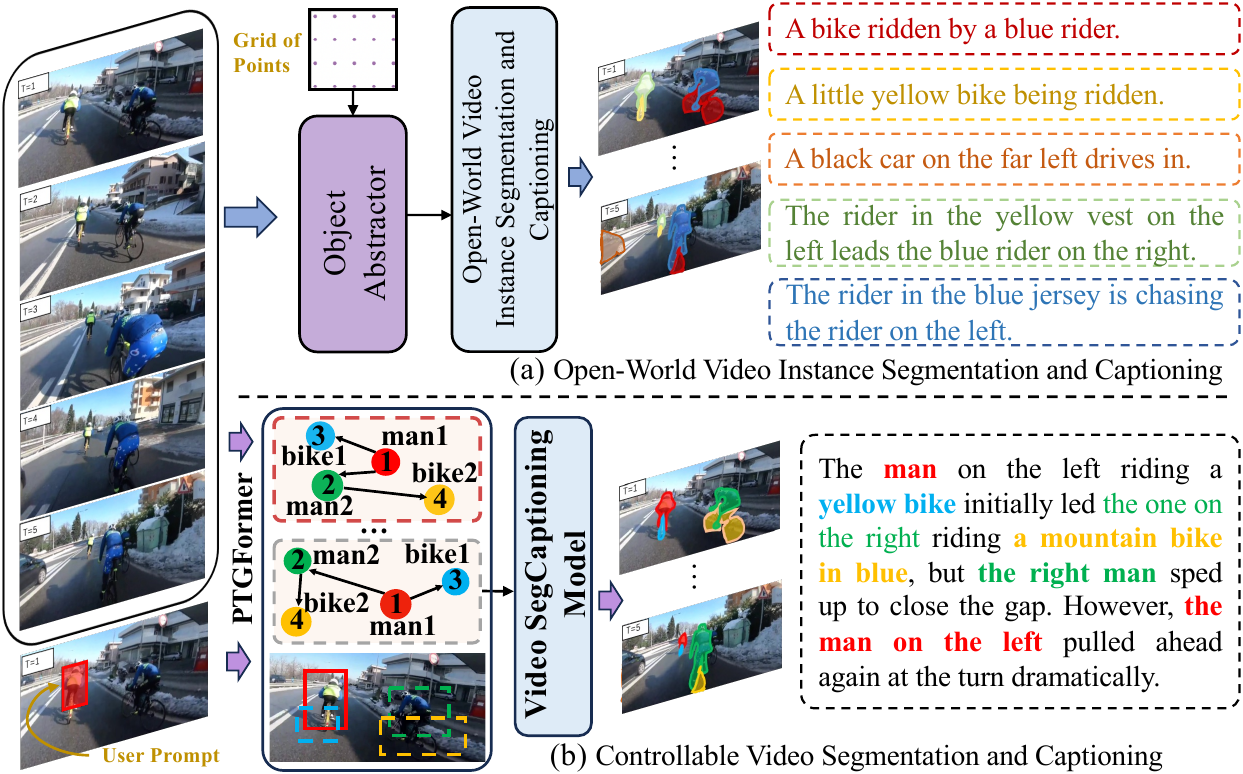}
	\caption{An example to illustrate the difference between the segmentation and captioning method and our approach, which allows users to provide a bounding box to generate multimodal outputs that are tailored to the user's intent.}
    \label{fig:motivation}
\end{figure}

\IEEEPARstart{V}{ideo} understanding \cite{he2024ma, song2024moviechat}, which entails the precise extraction of semantic information from temporal visual data, holds immense potential for applications in autonomous driving \cite{li2023bi, lin2024echotrack}, human-computer interaction \cite{lin2023click, zhang2024pvpuformer}, robots \cite{reddy2022first, yang2025learning}, and intelligent surveillance \cite{xia2020multi, fang2023surveillance}. 
Current video understanding tasks encompass video segmentation \cite{qi2022occluded, wu2022defense, zhu2024exploring, qiu2024visual, li2025visual}, action detection \cite{zhao2022real, chen2021watch}, and captioning \cite{wang2019controllable, liu2021o2na, zhang2025sgdiff}, among others. In these tasks, videos are transformed into semantic representations, such as masks, tags, and captions.

With the rapid advancement of large model architectures, particularly in their ability to represent modalities and perform cross-modal transformations, video understanding has transitioned from traditional single-task, unimodal outputs to multitask, multimodal outputs. These multimodal representations exhibit intricate interdependencies between modalities, effectively harnessed by multimodal large models to provide more comprehensive and nuanced feedback, thereby enhancing user comprehension. 
For instance, as depicted in \cref{fig:motivation} (a), \cite{choudhuri2024ow} associates each video instance with both a corresponding caption and mask, delivering a dual visual-textual interpretation of the video content.
While these innovations have undoubtedly expanded the informational breadth available to users for video comprehension, they still lack direct user interaction. Consequently, the generated outputs often contain superfluous information that may not fully align with specific user requirements, thereby limiting their utility in precise, context-driven applications.

To address this gap, this paper introduces a novel task,``Controllable Video Segmentation and Captioning'' (SegCaptioning), which empowers users to provide a visual prompt—such as a bounding box delineating an object of interest—to concurrently generate a caption and several corresponding masks across temporal frames, as depicted in \cref{fig:motivation} (b). Through this prompt-driven paradigm, the task challenges the model to autonomously identify all potential objects of interest and accurately model their spatio-temporal interrelationships, ensuring alignment with user-specific intent. Additionally, the multimodal outputs necessitate a precise correspondence between each mask and its respective caption word, posing substantial challenges in achieving robust cross-modal fine-grained alignment.

To this end, this paper introduces a pioneering ``Scene Graph guided Fine-grained SegCaptioning Transformer'' (SG-FSCFormer), which leverages structured scene graphs to translate user intent from a basic prompt into an intent subgraph which is used to guide the generation of correlated captions and masks for precise multimodal content understanding tailored to user needs. Specifically, given a video and a user prompt, such as a bounding box around an object, we first design a Prompt-guided Temporal Graph Former (PTGFormer) to identify objects of interest and their complex relationships, represented as a prompt-related graph feature. 
PTGFormer incorporates a novel prompt adaptor to remove irrelevant nodes and edges from the global scene graph by thoroughly exploring the spatial and temporal correlations between each object and the prompt object, resulting in a refined representation that aligns closely with user intent. Next, we design a Fine-grained Mask-linguistic Decoder to guide the generation of captions and masks. Concretely, the decoder first employs a Graph-guided Iterative Query Former that converts graph features into language embeddings, which are subsequently processed by a frozen large language model to predict captions. For mask prediction, the revised SAM2 decoder receives both images and graph features, and then simultaneously predicts masks associated with their position information appearing in the caption. This cross-modal fine-grained correspondence is supervised by a proposed fine-grained alignment loss. 
Furthermore, to achieve precise alignment between masks and caption words, the Mask-linguistic decoder employs a Cross-modal Multi-entity Contrastive loss, which draws the embeddings of positive caption words and their corresponding masks closer while distancing negative ones. Consequently, the improved cross-modal representations help the decoder to produce high-quality caption-mask pairs, enhancing user comprehension. 
Extensive experiments on two annotated datasets (``LV-VIS'' and ``OVIS'') demonstrate the effectiveness of SG-FSCFormer, achieving state-of-the-art performance surpassing existing methods. 
In summary, the main contributions of this work are as follows:
\begin{itemize}
    \item We introduce a pioneering task, ``Controllable Video Segmentation and Captioning'', marking the inaugural controllable video multimodal interpretation. 
    This task enables the flexible and nuanced multimodal interpretations of video content, meticulously tailored to align with specific user requirements.
    To support this task, we present two newly annotated datasets based on LV-VIS and OVIS, which will be publicly released to facilitate future research~\footnote{https://github.com/XuZhang1211/SG-FSCFormer}.
    \item We propose an innovative Scene Graph-guided Fine-grained SegCaptioning Transformer, featuring a Prompt-guided Temporal Graph Former that captures user intent by thoroughly exploring spatial and temporal correlations between objects and the prompt, yielding a precise prompt-related graph feature aligned with user intent.
    \item We devise a Mask-linguistic Decoder to collaboratively generate accurate, coherent multimodal outputs using a Multi-entity Contrastive loss. The outcomes achieve instance-level cross-modal alignment, significantly enhancing the user's comprehension of video content.  
\end{itemize}

\section{Related work}
\label{sec:related_work}

\subsection{Controllable Video Captioning} \label{sec:related_VU}
Recent advancements in vision-language models (VLMs) have significantly enhanced models' understanding of visual content, leading to substantial improvements in video captioning \cite{he2024ma, song2024moviechat, ren2024timechat}, which aims to describe global video content in natural language \cite{zhou2024streaming, zhou2023dense}, However, this global captioning paradigm fails to satisfy users' personalized requirements, motivating the development of controllable video captioning \cite{wang2019controllable, nitta2024fine} to specify visual elements guided by user prompts. 
For example, \cite{yuan2020controllable} uses an exemplar sentence to directly control syntax and style during captioning, while \cite{yao2024edit} introduces a video caption editing (VCE) task that automatically revises an existing video description based on user requests. Beyond text prompts, \cite{teng2023sovc} enables users to specify visual bounding boxes for subject-focused descriptions. To achieve fine-grained control over caption length or topic, \cite{nitta2024fine} employs multi-hot vector representations to precisely regulate caption length, while \cite{liu2021o2na} proposes a topic-guided model to generate topic-oriented descriptions. 
In contrast, our method jointly generates captions and masks with precise alignment, delivering more meaningful and user-tailored video understanding.

\subsection{Referring Video Segmentation} \label{sec:related_VOC}
Referring video object segmentation (R-VOS) is a specialized subfield of video segmentation that seeks to localize and segment target objects across all video frames given a user-provided prompt \cite{he2024decoupling, li2024univs, miao2024temporally}. 
Classical video segmentation focuses on segmenting, tracking, and classifying objects drawn from a fixed set of training categories \cite{huang2022minvis, wu2022defense, zhang2025dvis++}. Recent efforts have moved beyond this closed-set assumption toward open-vocabulary video instance segmentation (OV-VIS) \cite{guo2023openvis, wang2023towards, fang2024unified, guo2025videosam}, which leverages the visual–language alignment capabilities of vision–language models (VLMs) \cite{radford2021learning}. 
For example, OV2Seg \cite{wang2023towards}, trained on image datasets, can be directly applied to videos at test time to recognize novel categories, while OVFormer \cite{fang2024unified} improves generalization via a lightweight module that aligns query embeddings with CLIP image embeddings, narrowing the domain gap. 
Although OV-VIS alleviates the limitations of predefined taxonomies, it still does not guarantee segmentation that is tailored to diverse user prompts and intent.
R-VOS \cite{wu2022language, xiao2024temporal, yang2024language, mei2024slvp} addresses this gap by conditioning segmentation on natural-language (or other) prompts; however, effectively modeling object dynamics and long-range temporal correlations remains challenging. 
To address this, \cite{tang2023temporal} introduces a temporal collection-distribution mechanism that facilitates interactions between the reference token and object queries, while \cite{he2024decoupling} suggests decoupling video-level referring expressions into static and motion components to enhance temporal comprehension. 
To adapt to novel scenes, \cite{li2023learning} proposes an effective few-shot R-VOS model to enable rapid semantic learning and adaptation across diverse scenarios. \cite{zhu2024exploring} explores the use of a pretrained text-to-video (T2V) diffusion model, incorporating specially designed components for R-VOS. 
Besides text prompts, SAM2 \cite{ravi2024sam} first introduces an interactive video segmentation method built on SAM \cite{kirillov2023segment}, using visual prompts for high efficiency. 
In contrast, our method not only provides open-vocabulary labels for the predicted masks but also supports the automatic expansion of a concise user prompt into multiple semantically related objects, all organized by a semantic caption with fine-grained cross-modal alignment.

\subsection{Video Multimodal Interpretation} \label{sec:related_VISC}
Advancements in multimodal large models \cite{radford2021learning, li2022blip, zhu2023minigpt, maaz2023video, lai2024lisa} have driven the development of video multimodal interpretation, which aims to generate correlated multimodal outputs for enhanced video understanding. These approaches can be broadly categorized into two primary paradigms. Box-caption methods, such as \cite{zhou2023dense}, integrate object detection, tracking, and trajectory captioning within videos by associating instance-level bounding boxes with corresponding captions. 
Expanding on this idea, controllable box-caption generation incorporates user guidance to improve interpretability. For example, SMOTer \cite{li2025beyond} introduces dynamic conditional inputs to guide object trajectory generation and produce interaction-aware captions, enabling user-specified tracking granularity. 
Another line of research focuses on mask-caption pair generation, where methods like OW-VisCaptor \cite{choudhuri2024ow} perform segmentation, tracking, and captioning using open-world object queries, producing instance masks paired with captions for each detected object. 
However, these approaches describe each identified target in isolation, overlooking the surrounding contextual information and often resulting in incomplete or fragmented descriptions.
VideoGLaMM~\cite{munasinghe2025videoglamm} generates broad and unfocused captions without precisely addressing the user’s region of interest. At the same time, it relies solely on text-level referring features for mask decoding and processes each target’s textual features independently within the segmentation module. 
This design not only amplifies inherent cross-modal discrepancies between textual and visual modalities but also aggravates ambiguities among visually similar objects, ultimately hindering accurate pixel-level segmentation.

Building on these limitations, we introduce a novel Controllable Video Segmentation and Captioning task that leverages user-specified prompts to produce correlated captions and segmentation masks. 
Unlike prior approaches, our framework emphasizes faithfully capturing user intent by inferring related object masks guided by a prompt box. 
Furthermore, it effectively bridges dynamic visual contexts with textual descriptions, enabling the collaborative generation of multimodal outputs that mitigate cumulative errors and promote complementary information exchange across modalities. Importantly, our fine-grained alignment mechanism ensures that noun entities in the generated captions are tightly and accurately grounded to their corresponding segmented objects, thereby achieving precise cross-modal consistency.

\section{Method}
\label{sec:formatting}

\begin{figure*}[ht!]
	\centering
	\includegraphics[width=1\linewidth]{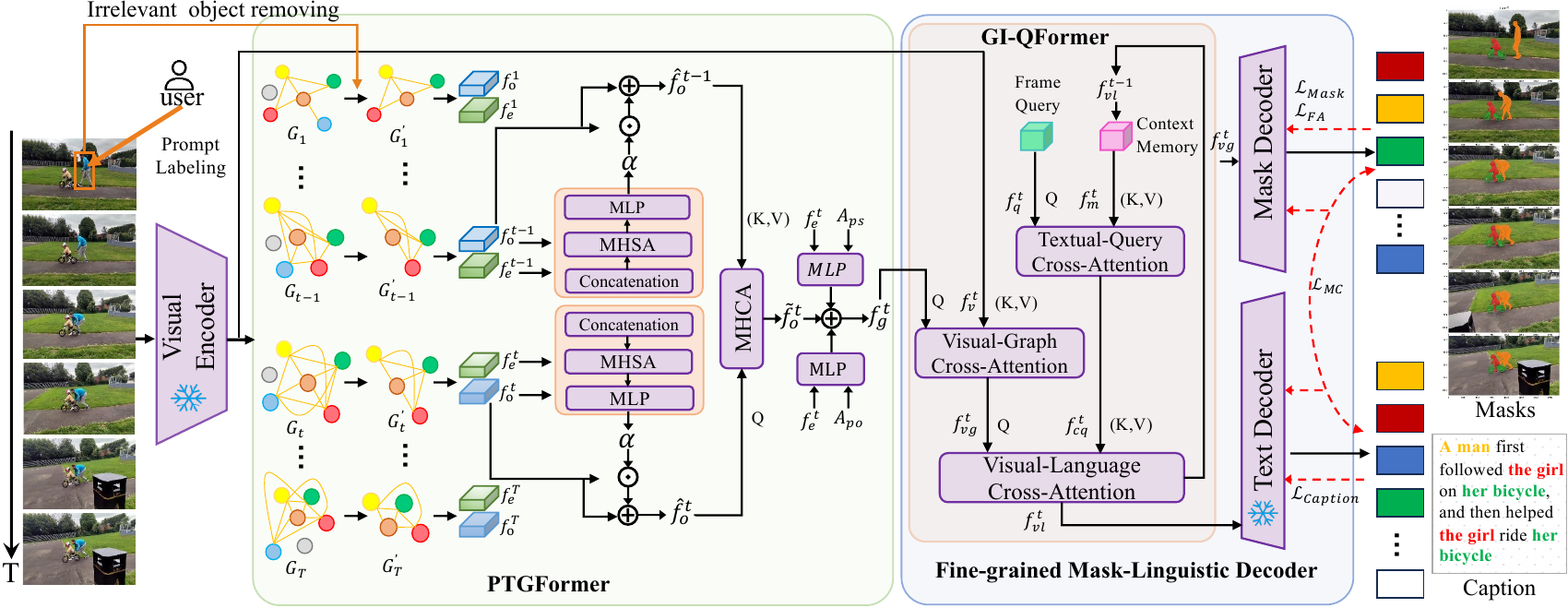}
	\caption{Framework of our Scene Graph-guided Fine-grained SegCaptioning Transformer (SG-FSCFormer), consisting of three components:
a Visual Encoder, a Prompt-guided Temporal Graph Former (PTGFormer), and a Fine-grained Mask-Linguistic Decoder (MLDecoder), where the mathematical symbols are explained in the corresponding modules.}
    \label{fig:networks}
\end{figure*}

\subsection{Overview}
Given a video $V$ and a user prompt $P_o$, such as a bounding box around an object $o$ in the first frame, controllable video SegCaptioning extracts user-specified semantic information from the video according to $P_o$, which is represented as a caption sentence $S$ paired with related masks $M$ that cover all instances in $S$. 
We propose a novel ``\textbf{S}cene \textbf{G}raph guided Fine-grained \textbf{S}eg\textbf{C}aptioning Trans\textbf{former}'' (SG-FSCFormer), designed to simultaneously generate a caption with semantically aligned masks, which consists of three main components (see~\Cref{fig:networks}): a visual encoder, a Prompt-guided Temporal Graph Former (PTGFormer), and a Fine-grained Mask-Linguistic Decoder. 
The visual encoder extracts visual features for all video frames. On this basis, the PTGFormer generates temporal scene graph features to capture objects of interest and their complex relationships based on the input prompt, ensuring an accurate representation of user intent. 
The fine-grained mask-linguistic decoder transforms these graph features along with the encoded visual features into paired masks and captions. Notably, this collaborative decoding on both modalities is supervised by the Cross-temporal Fine-grained Alignment Loss, which aims to capture the complex fine-grained correspondence between each mask and captions words in instance-level instead of class-level, providing detailed cross-modal semantic relationships to facilitate users' understanding.  

\subsection{Prompt-guided Temporal Graph Former}
Given a video $V \in \mathbb{R}^{T \times H \times W \times 3}$ with height $H$, width $W$, and $T$ frames, our PTGFormer generates temporal scene graph features to capture the visual content of interest to users across the frames, as well as the complex semantic context embedded within them. Specifically, the PTGFormer first generates global scene graphs $\{G_t\}_{t=1}^{T}=\{(N_t, E_t)\}_{t=1}^{T}$ 
for each frame $F_t$ following \cite{zellers2018neural}, where $N_t$ and $E_t$ denote the sets of nodes and edges, respectively. To filter out redundant nodes and edges in $G_t$ that are not relevant to the user’s interest, PTGFormer aims to identify a subgraph $G'_t \subseteq G_t$, guided by the prompt $P_o$, to well align with the mask-caption pair 
$(M, S)$ for model training. The goal of this selection is to maximize the expression below:
\begin{equation}
P(G'_t\overset{s}=(S, M)|G_t,P_o).
\end{equation}
where $\overset{s}=$ indicates the semantic consistency between $G'_t$ and $(M, S)$, approximated by object consistency between $N_t$ of $G'_t$ and $(M, S)$. To achieve this, 
we design an adaptive prompt adaptor to automatically extract a subgraph $G'_t$ from $G_t$. Concretely, the adaptor first removes nodes and edges in $G_t$ that are not connected to the prompt object, creating a coarse subgraph with node features $f_{o}^t \in \mathbb{R}^{L_o\times D}$ and edge features $f^t_e \in \mathbb{R}^{L_e \times D}$, 
where $L_o$, $L_e$ represent the number of nodes and edges, and $D$ is the feature dimension. Next, the adaptor concatenates $f_{o}^t$ and $f^t_e$ and injects them into several self-attention blocks by exploring the complex correlations among nodes, followed by a mapping layer that outputs response scores $\alpha \in \mathbb{R}^{L_o}$, expressed as follows:
\begin{align}
\alpha = \sigma \left(\text{MHSA}([f_{o}^t;f_{e}^t] \right)), \label{eq:2}
\end{align}
where $\text{MHSA}$ denotes the multi-head self-attention operation, $\sigma$ represents a mapping layer implemented via an MLP, and $\alpha$ reflects the association strength between each node and the prompt object. Since the edge feature $f^t_e$ is centered on the prompt object, the deep exploration of both node and edge features allows us to effectively predict the association strength between each node and the prompt object. We then update the node features by incorporating this association strength, which helps emphasize the highly relevant nodes:
\begin{align}
\hat{f}_o^t = (1 + \alpha) \cdot f_{o}^t, 
\label{eq:3}
\end{align}
where $\hat{f}_o^t \in \mathbb{R}^{L_o\times D}$ represents the prompt-guided reinforcement features of nodes that account for spatial correlations between each node and the prompt object. Additionally, the adaptor further incorporates temporal correlations between neighboring frames to update the node features:
\begin{equation}
\tilde{f}^t_{o} = \text{MHCA}(\hat{f}^{t}_{o}, \hat{f}^{t-1}_{o}, \hat{f}^{t-1}_{o}),
\label{eq:5}
\end{equation}
where $\text{MHCA}$ represents the multi-head cross-attention layers. By modeling spatio-temporal scene graphs, the enhanced feature $\tilde{f}^t_{o} \in \mathbb{R}^{L_o\times D}$ effectively captures the user’s intent and the dynamic semantic information across frames, providing more accurate features for subsequent steps.

Finally, we combine both the node feature $\tilde{f}^t_{o}$ and edge feature $f^t_e$ to generate the final graph feature. 
Specifically, the edge feature $f^t_e$ is first multiplied by two adjacency matrices, $A_{ps}$ and $A_{po}$ from the graph $G'_t$, which capture predicate-subject and predicate-object relationships, respectively. The edge feature is then updated using two fully connected layers, $\text{MLP}_{ps}$ and $\text{MLP}_{po}$, to ensure size consistency with $\tilde{f}^t_{o}$. As a result, the final prompted-related graph feature $f^t_g \in \mathbb{R}^{L\times D}$ is calculated as:
\begin{equation}
f^t_g = \tilde{f}^t_{o} + \text{MLP}_{ps}(f^t_e A_{ps}) + \text{MLP}_{po}(f^t_e A_{po}).
\label{eq6}
\end{equation}

Unlike typical scene graph features, the generated $f^t_g$ has the following key characteristics: First, it effectively filters out irrelevant nodes and edges based on the prompt object, thereby better capturing the user’s intent. Second, the adaptor thoroughly explores the correlations between each graph node and the prompt object, as well as temporal correlations across frames. This enables the model to identify notable objects, which is crucial for making accurate caption predictions.
Moreover, it enhances the model’s ability to re-identify the target when it is occluded or temporarily disappears in intermediate frames. 

\subsection{Fine-grained Mask-linguistic Decoder} 
The Fine-grained Mask-Linguistic Decoder is designed to generate mask regions that highlight user-specified areas of interest, along with an associated caption that semantically aligns with the predicted masks.
Moreover, the decoding outcomes specify the correspondence between each mask and individual caption words, ensuring fine-grained alignment.
Technically, our decoder first leverages scene graph features to guide the feature generation process, integrating both textual and visual-semantic information through a Graph-guided Iterative Query Former. Subsequently, these features are decoded to produce the corresponding caption and masks.

\subsubsection{Graph-guided Iterative Query Former}
To effectively generate features that integrate textual and visual semantics, we employ a Graphs-guided Iterative Query Former (GI-QFormer), as~\Cref{fig:networks} shows. Given a scene graph feature $f^t_{g} \in \mathbb{R}^{L \times D}$, an image feature $f_v^t \in \mathbb{R}^{U \times D}$ with size $U$, and a learnable language query $f_q^t \in \mathbb{R}^{L \times D}$ at frame $t$, our GI-QFormer generates language embeddings that align with user intent through three multi-head cross-attention layers: visual-graph cross-attention, textual-query cross-attention, and visual-language cross-attention. In the visual-graph cross-attention layer, $f^t_{g}$ serves as the query, and $f_v^t$ acts as both the key and value to obtain the prompt-related visual feature $f^t_{vg} \in \mathbb{R}^{L \times D}$, which accurately reflects the user's demand guided by $f^t_{g}$.
In the textual-query cross-attention layer, a learnable embedding $f^t_q$ is used as the text query, while a context memory embedding  $f^t_{m} \in \mathbb{R}^{K \times D}$ serves as the key and value to generate a context-related language feature $f^t_{cq} \in \mathbb{R}^{L \times D}$, where $K$ is the number of compressed temporal frames. The initial $f^t_{m}$ is randomly set and updated iteratively by \cite{zhou2024streaming} for long-term video efficiency, based on the outputs received from GI-QFormer at each time step. Finally, in the visual-language cross-attention layer, $f^t_{vg}$ is used as the query, and $f^t_{cq}$ serves as the key and value to generate the prompt-related textual feature  $f^t_{vl} \in \mathbb{R}^{L \times D}$. Here, $f^t_{vg}$ acts as a user-interested visual query, dynamically retrieving relevant textual information from $f^t_{cq}$. Consequently, the generated $f^t_{vl}$ captures the user’s intent for caption generation. All the three cross-attention layers are iteratively applied to generate the final text embedding $f^T_{vl}$ for caption generation.

\begin{figure}[t!]
	\centering
	\includegraphics[width=1 \linewidth]{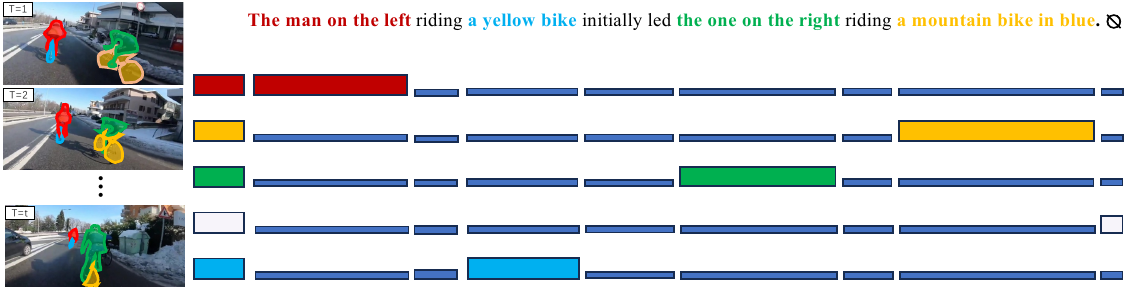}
	\caption{Prediction by the revised decoder which is able to output the position of caption words for each mask instance.}
    \label{fig_sfloss}
\end{figure}

\subsubsection{Decoding} 
For caption generation, we employ a frozen large language model (LLM)~\cite{chiang2023vicuna} as the text decoder. The LLM takes $f^T_{vl}$ as input to generate a caption. The caption loss $\mathcal{L_{\text{caption}}}$ is computed as Cross-Entropy (CE) loss, as detailed in \cite{luo2023semantic}.
For mask generation, we employ the SAM2 decoder~\cite{ravi2024sam} to generate masks by employing the original mask loss $\mathcal{L_{\text{Mask}}}$. In addition, we add an branch consisting of a two-layer MLP to predict a referring probability vector, which indicates the object location of each mask in the caption, as illustrated in~\Cref{fig_sfloss}. Specifically, 
the mask decoder takes the $f^t_{vg}$ as input and outputs binary region masks $M^t \in \mathbb{R}^{N\times H\times W}$, along with a probability distribution matrix $V^t \in \mathbb{R}^{N \times L_s}$, where $N$ denotes the number of objects and $L_s$ represents the caption length.  
We introduce a fine-grained alignment loss $\mathcal{L}_{\text{FA}}$ to train the model for accurately predicting $V^t$:
\begin{equation}
\mathcal{L}^t_{\text{FA}} = \text{BCE}(V^t, Y) , 
\label{md}
\end{equation}
where $Y$ represents the ground-truth of the probability distribution matrix labeled in our experiments.

To collaboratively predict both segmentation and captioning results, our approach explicitly align cross-modal features between each mask with its corresponding caption words by using a Multi-entity Contrastive loss $\mathcal{L}_{\text{MC}}$. 
Given a predicted mask-caption pair $(M, S)$, where $M$ consists of $T$ sets $\{M_t\}_{t=1}^T$  for $T$ frames, $\mathcal{L}_{\text{MC}}$ measures the correspondence between masks in $M_t$ and words in $S$ within each static frame $F_t$. Notably, there is no strict one-to-one correspondence between masks and words. For instance, the word ``elephant'' may correspond to multiple masks, and conversely, a single mask may relate to multiple distinct words. To account for this, we aim for the $i$-th mask embedding 
$\mathbf{m}_i^t$ to be close to the set of positive word embeddings ${\mathbf{s}_i }^+$, while remaining distant from unrelated word embeddings $\mathbf{s}_j$ . Similarly, we expect the $i$-th word embedding $\mathbf{s}_i$ to be similar to the set of positive mask embeddings $\{\mathbf{m}_i^t\}^+$ and dissimilar to irrelevant embeddings $\mathbf{m}_j^t$.
Consequently, $\mathcal{L}_{\text{MC}}$ is defined as:
\begin{small}
\begin{equation}
\begin{split}
\mathcal{L}^t_{\text{MC}} = &\!- \sum_{i=1}^{|M_t|} \frac{1}{\vert \{ \mathbf{s}_i \}^+ \vert } \sum_{\mathbf{s}_i \in  \{ \mathbf{s}_i \}^+ } \!\!\!\!\!\log \frac{\exp(\mathbf{m}_i^{t\top}  \mathbf{s}_i / \tau)}{\sum\limits_{j \neq i} \exp(\mathbf{m}_i^{t\top} \mathbf{s}_j/\tau) } \\
&- \!\!\sum_{i=1}^{|S|} \frac{1}{\vert \{ \mathbf{m}_i^t \}^+ \vert} \sum_{\mathbf{m}_i^t \in \{ \mathbf{m}_i^t \}^+} \!\!\!\!\!\log \frac{\exp( \mathbf{s}_i^\top  \mathbf{m}_i^t / \tau)}{\sum\limits_{j \neq i} \exp(\mathbf{s}_i^\top  \mathbf{m}_j^t /\tau) } ,
\end{split}
\end{equation}
\end{small}
where $\tau$ is a learnable parameter, $|M_t|$ and $|S|$ denote the number of mask embeddings and word embeddings at the $t$-th frame, respectively. Finally, the overall loss of our mask-linguistic decoder is composed of three parts by using a balancing weight ${\lambda}$: the captioning loss $\mathcal{L}_\text{Caption}$, the segmentation loss $\mathcal{L}_{\text{Mask}}+\mathcal{L}_{\text{FA}}$, and the cross-modal Multi-entity Contrastive loss  $\mathcal{L}_{\text{MC}}$, which is expressed as:
\begin{equation}
  \mathcal{L}_{\mathrm{total}} = \mathcal{L_{\text{Caption}}} + (\mathcal{L_{\text{Mask}}} + \mathcal{L_{\text{FA}}}) + {\lambda} \mathcal{L_{\text{MC}}}.
\label{eq:total_loss}
\end{equation}

\begin{figure}[t!]
	\centering
\includegraphics[width=0.99\linewidth]{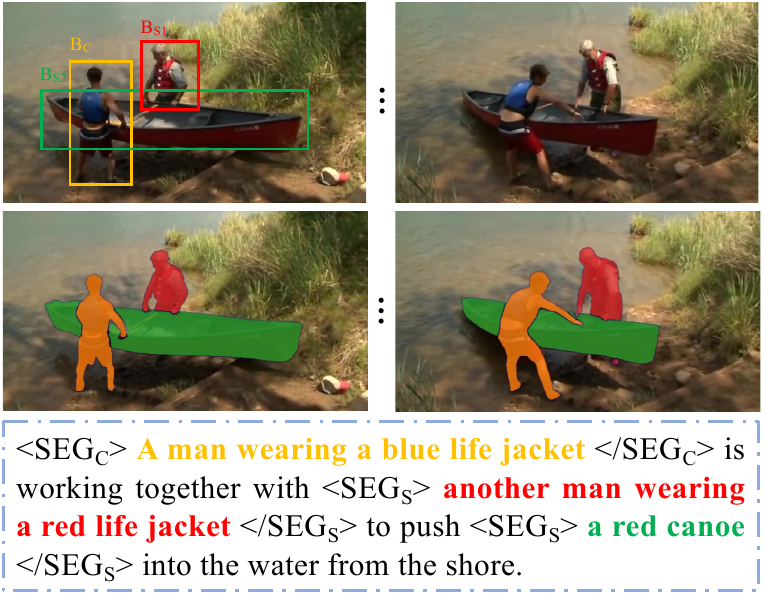}
	\caption{An example illustrating our data annotation.}
	\label{fig:data_anno}
\end{figure}

\section{Dataset Annotation}\label{app:dataset_annotation}
We augment existing video segmentation datasets with textual annotations to facilitate our SegCaptioning task. 
Specifically, given an input video, the original dataset provides ground-truth masks and category labels for each target.
We first compute the enclosing rectangle around each mask to obtain its bounding box.
Annotators then review these boxes across frames and provide a caption about the bounding target.
In each caption, the bounding box corresponding to the subject is designated as the central box $B_c$, which serves as the user prompt box during training, while all other relevant boxes are treated as association boxes $B_s$.
To ensure annotation quality, two independent annotators annotate each video, and their results are cross-checked for consistency. 

For a single video, multiple prompt box–caption pairs can be created for different objects of interest.
For example, in a given video, the central box round ``a man wearing a blue life jacket" (see~\cref{fig:data_anno}).
The required annotations are as follows: 
Central box $\to$ ``a man wearing a blue life jacket", Association box 1 $\to$ ``another man wearing a red life jacket", Association box 2 $\to$ ``a red canoe".
To structure the caption, we use $\langle \text{SEG}_C\rangle \langle /\text{SEG}_C\rangle$ to enclose nouns corresponding to the central box and $\langle \text{SEG}_{S}\rangle \langle /\text{SEG}_{S}\rangle$ for nouns corresponding to the association box. The final annotated caption is:
$\langle \text{SEG}_C\rangle$A man wearing a blue life jacket$\langle /\text{SEG}_C\rangle$ is working together with $\langle \text{SEG}_{S}\rangle$another man wearing a red life jacket$\langle /\text{SEG}_{S}\rangle$ to push $\langle \text{SEG}_{S}\rangle$a red canoe$\langle /\text{SEG}_{S}\rangle$ into the water from the shore.

During model's training and inference, only the central box is provided as the user prompt. A temporal scene graph is then used to adaptively identify other objects associated with the target (with supervision from association box labels during training), which are subsequently used for video segmentation and caption generation.
Additionally, to ensure that the target of interest appears in the first frame of each sample, we filter out instances where the prompted target is absent in the first frame and discard all preceding frames. This preprocessing step ensures dataset consistency and maintains a clear focus on the target from the beginning.
Our work is the pioneering study to implement the fine-grained alignment between caption tokens and masks guided by visual prompts, which inevitably introduces new annotated data. To support future research, we will publicly release the constructed dataset.

\textbf{Discussing annotation quality, and potential bias.}
While manual annotation may affect scalability, we have taken several measures to ensure annotation quality.
The original dataset provides precise masks from which our bounding boxes are generated, ensuring exact localization and eliminate potential localization bias.
Captions are generated by two independent annotators per video, refined by LLMs for grammatical correctness, and then cross-checked to ensure consistency and high inter-annotator agreement, which guarantees annotation quality. 
To better reflect diverse user intentions and reduce perspective bias, we allow multiple valid central boxes within each scene. 
When different central boxes are designated, each team may produce distinct captions to reflect diverse user perspectives and intent.
Instead of merging these into a single consensus caption, we preserve the diversity across annotations. This approach enables the dataset to better represent nuanced shifts in user intent and provides richer supervisory signals for models aimed at relation-aware and context-sensitive understanding.

\section{Experiments}  \label{experiments}

\begin{figure*}[ht!]
	\centering
\includegraphics[width=1\linewidth]{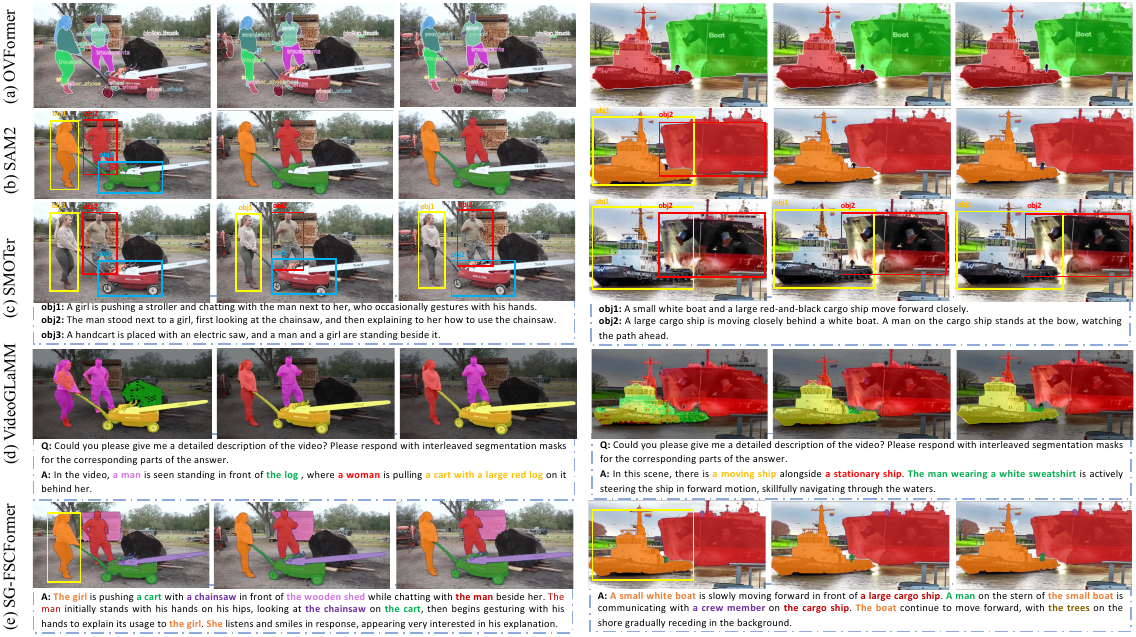}
	\caption{Quantitative results by different approaches tested on the LV-VIS and OVIS datasets, where the color indicates the matching relationship between masks and words.}
	\label{fig:results}
\end{figure*}

\subsection{Datasets and Metrics}
We conduct our experiments on two video instance segmentation datasets, ``LV-VIS'' \cite{wang2023towards} and ``OVIS'' \cite{qi2022occluded}, by incorporating new prompt and caption annotations.

\textbf{LV-VIS} is a large vocabulary video instance segmentation dataset consisting of $4,828$ real-world videos across $1,196$ categories. The dataset includes annotated masks and is divided into three subsets: $3,083$ for training, $837$ for validation, and $908$ for testing. We have enriched this dataset by adding (box, caption) pairs. Specifically, several annotators were instructed to label one of the masked targets in each video using a bounding box in the first frame and then to describe an event centered on this target in relation to other masked objects. 
To provide fine-grained alignment for model training, all the masked objects were labeled with the position information to reflect their appearance in the captions. 
As a result, each video contains $1$-$3$ bounding boxes with corresponding captions. This augmentation yields a total of $9,588$ (box, caption) pairs, averaging approximately $1.98$ (box, caption) pairs per video.

\textbf{OVIS} is a dataset designed for occluded video instance segmentation, which requires detecting, segmenting, and tracking instances in scenes with significant occlusions. 
The dataset consists of $901$ videos across $25$ object categories, divided into $607$ training, $140$ validations, and $154$ testing videos.
We have similarly extended this dataset to include (box, caption) pairs for each video, ensuring alignment between caption words and their visual targets.
In total, we have added $2,190$ (box, caption) pairs, averaging approximately $2.4$ (box, caption) pairs per video.

We evaluate our model from three perspectives: 
the quality of caption generation, the quality of video segmentation, and the quality of the alignment between caption words and masks.
Caption quality is assessed using METEOR, SPICE, and CIDEr \cite{anderson2016spice,vedantam2015cider}, while
video segmentation accuracy is evaluated using the $J\&F$ \cite{perazzi2016benchmark}.
The alignment is evaluated at both class-level and instance-level. 
For class-level alignment, we use average precision (AP) to measure the classification accuracy of masks. 
For instance-level alignment, our approach first finds the most relevant phrase for each mask predicted by the segmentation decoder. Then, it calculates the textual similarity between the phrase and the caption words in the ground truth. If the similarity exceeds $0.5$, the prediction is considered correct. We then calculate the average precision on all the instances to obtain the instance-level mAP.

\subsection{Implementation Details} 
We use SwinB~\cite{liu2021swin} as our visual encoder, and the SAM2 decoder~\cite{ravi2024sam} and Vicuna-7B~\cite{chiang2023vicuna} as the decoders for mask and caption generation, respectively. In the PTGFormer, the graph feature $f^t_g$ is configured with $L=100$ and $D=512$, while the context memory has a memory length of $K=10$.
For the total loss function in Eq.~\eqref{eq:total_loss}, we set $\lambda=2$ to balance the model's optimization. 
During training, we apply horizontal flip augmentation and resize each image to a square size of $1024 \times 1024$. 
We optimize the model using the AdamW optimizer with an initial learning rate of $5.0 \times 10^{-4}$ and train for $20$ epochs with a batch size of $1$ per GPU. All experiments are conducted on $4$ A6000 GPUs.

\begin{table*}[t!]
    \centering
    \caption{Performance comparison with the advanced video captioning methods. where the results on the LV-VIS and OVIS are reproduced and that on the YouCook2 is 
 cited from the references.}
    \label{tab_1}
    \renewcommand\arraystretch{1.0}
\centering
\setlength\tabcolsep{1.0mm}
\resizebox{1\linewidth}{!}{
\begin{tabular}{l|c|ccc|ccc|ccc}
\toprule
\multirow{2}{*}{Model} & \multirow{2}{*}{LLM} & \multicolumn{3}{c|}{LV-VIS} & \multicolumn{3}{c|}{OVIS} & \multicolumn{3}{c}{YouCook2} \\
                        & & METEOR & SPICE & CIDEr & METEOR & SPICE & CIDEr & METEOR & SPICE & CIDEr \\ \midrule
Vid2Seq  $_{\mathrm{CVPR23}}$\cite{yang2023vid2seq} & T5-Base & 16.3 & 24.7   & 107.4   & 18.0   & 33.0  & 106.8 & 9.3   & 7.9  & 47.1  \\
SMOTer $_{\mathrm{ECCV24}}$\cite{li2025beyond} & GRiT & 16.9   & 25.0   & 109.6 & 17.8   & 32.8  & 108.1 & -   & -  & - \\
MA-LMM $_{\mathrm{CVPR24}}$\cite{he2024ma}  & Vicuna-7B & 17.1   & 25.2   & 110.6 & 18.6   & 33.1  & 109.3 & 17.6   & 31.5  & 131.2 \\
VideoGLaMM $_{\mathrm{CVPR25}}$\cite{munasinghe2025videoglamm}  & Phi3-Mini-3.8B & 16.7   & 25.7   & 112.0 & 19.5   & 33.9  & 112.8 & 15.4   & 28.7  & 124.3 \\
SG-FSCFormer (ours) & Vicuna-7B & \textbf{19.3}   & \textbf{26.8}   & \textbf{112.5} & \textbf{21.4}   & \textbf{35.2}  & \textbf{113.7} & \textbf{20.5}   & \textbf{33.4}  & \textbf{139.6} \\ 
\bottomrule
\end{tabular}
}
\end{table*}

\subsection{Comparison with state-of-the-art methods}
We compare SG-FSCFormer with several state-of-the-art methods in video captioning, video segmentation, and multimodal video interpretation. 
The results are summarized in~\cref{tab_1}, \cref{tab_2}, and \cref{tab_3}, respectively.
Note that “-” indicates that the result is not reported by the original paper or the model is unavailable.

\begin{table}[t!]
    \centering
    \caption{Performance comparison between our approach and the two controllable video segmentation methods.}
    \label{tab_2}
    \renewcommand\arraystretch{1.0}
\centering
\setlength\tabcolsep{1.0mm}
\resizebox{1\linewidth}{!}{
\begin{tabular}{l|ccc|ccc}
\toprule
\multirow{2}{*}{Model} & \multicolumn{3}{c|}{LV-VIS} & \multicolumn{3}{c}{OVIS} \\
                         & {${J}\&{F}$} & ${J}$ & ${F}$ & {${J}\&{F}$} & ${J}$ & ${F}$ \\ \midrule
SAMURAI $_{\mathrm{arXiv24}}$\cite{yang2024samurai} & 87.3 & 85.6 & 89.1  & 72.0  & 69.2 & 74.9   \\
SAM2 $_{\mathrm{ICLR25}}$\cite{ravi2024sam} & 85.6 & 83.3 & 87.9   & 68.7  & 66.0 & 71.3  \\ 
SG-FSCFormer (ours) & \textbf{87.8} & \textbf{85.9} & \textbf{89.7}   & \textbf{74.6}  & \textbf{72.5} & \textbf{76.7}  \\
\bottomrule
\end{tabular}
}

\end{table}

\textbf{Video Captioning.}
On our LV-VIS and OVIS datasets, all the captioning models are re-executed. As \cref{tab_1} shows, 
SG-FSCFormer achieves the best performance on all the metrics. 
Compared with approaches that generate global descriptions of entire scenes (e.g., Vid2Seq, MA-LMM, and VideoGLaMM), which tend to generate broad and unfocused captions without precisely addressing the user’s region of interest, our method provides more targeted and semantically faithful outputs. 
Similarly, while SMOTer generates captions for individual objects, it neglects the surrounding contextual information, often leading to incomplete or fragmented descriptions. 
In contrast, SG-FSCFormer integrates both fine-grained object details and precise contextual cues through the user-guided visual prompt mechanism. By leveraging the proposed PTGFormer, the model proficiently translates a prompt into pertinent feature representations, thereby yielding accurate caption results that are closely aligned with user intent.
To further demonstrate the generalization capability of SG-FSCFormer, we extend our evaluation to the YouCook2 dataset \cite{zhou2018towards}, which contains pairs of query phrases and corresponding ground-truth descriptions for each video, and compare our results with the existing official results. 
For each video, we select the first occurrence of a ground-truth label mentioned in the description of the initial frame as the visual prompt for our method. 
Under this setting, our approach consistently achieves the best performance on all the metrics, further validating its robustness and adaptability.

\textbf{Controllable Video Segmentation.}
This experiment compares our approach with the controllable video segmentation methods. 
Given a video, users label a box from the target masks in the first frame for segmentation evaluation.
As \cref{tab_2} shows, our approach demonstrates commendable segmentation performance with the highest ${J}\&{F}$ scores of $87.8$ and $74.6$ on the LV-VIS and OVIS datasets, respectively. We guess this improvement stems from the collaborative decoding and the multi-entity contrastive loss, which could extract robust visual features from mask prediction.
In the OVIS dataset, since only the training set contains offline segmentation annotations, we reserved $100$ non-overlapping training videos for validation and used the rest for training.

\begin{table}[t!]
    \centering
    
    \caption{
    Performance comparison with several advanced video multimodal interpretation methods, where $^{\dag}$ denotes the evaluation using the instance-level AP.}
    \label{tab_3}
    \renewcommand\arraystretch{1.0}
\centering
\setlength\tabcolsep{0.2mm}
\resizebox{1\linewidth}{!}{
\begin{tabular}{l|c|ccc|ccc}
\toprule
\multirow{2}{*}{Model} & \multirow{2}{*}{Mask Decoder} & \multicolumn{3}{c|}{LV-VIS} & \multicolumn{3}{c}{OVIS} \\
                         & & AP &  AP\textsubscript{50} &  AP\textsubscript{75} & AP &  AP\textsubscript{50} &  AP\textsubscript{75} \\ \midrule
GLEE $_{\mathrm{CVPR24}}$\cite{wu2024general} &  MaskDINO & 23.9 & 24.6 & 23.3  & 27.1 & 45.4  & 26.3   \\
OVFormer $_{\mathrm{ECCV24}}$\cite{fang2024unified} & Mask2Former & 24.7& 31.1 & 26.5  & 21.3 & 38.5 & 20.8   \\
OW-VISCap $_{\mathrm{NIPS24}}$\cite{choudhuri2024ow} & Mask2Former & - & - & -   & 25.4   & 48.8 & 22.8    \\
SG-FSCFormer (ours) & Mask2Former & 25.2 & 31.8 & 27.5  & 27.8 & 49.0 & 27.2   \\
\hline
VideoGLaMM $_{\mathrm{CVPR25}}$\cite{munasinghe2025videoglamm} & SAM2 & 25.4& 32.2 & 27.6  & 28.0   & 49.3 & 26.9    \\
SG-FSCFormer$^{\dag}$ (ours) & SAM2 & 24.8 & 31.5 & 26.9  & 27.5 & 48.6 & 26.7   \\
SG-FSCFormer (ours) & SAM2 & \textbf{26.0} & \textbf{33.6} & \textbf{28.3}   & \textbf{28.7}  & \textbf{51.4} & \textbf{29.1}  \\ 
\bottomrule
\end{tabular}
}

\end{table}

\begin{table}[t!]
    \centering
    \caption{Performance comparison of PTGFormer using the adaptor with different components tested on the LV-VIS validation set, where ``spa.'' and ``tem.'' represent the use of spatial and temporal correlations, respectively.}
    \label{tab_4}
    \centering
\renewcommand\arraystretch{1.0}
\setlength\tabcolsep{2.8mm}
\resizebox{1\linewidth}{!}{
\begin{tabular}{cc|cc|c|c}
\hline 
Spa.    &   Tem.  & SPICE & CIDEr & {${J}\&{F}$} & AP \\ \midrule
-            & -              & 24.6   & 108.3 & 85.0  & 23.9      \\
$\checkmark$ & -              & 25.7   & 111.2 & 86.5   & 25.4      \\
-            & $\checkmark$   & 25.3   & 109.8 & 86.4   & 25.1      \\ 
$\checkmark$ & $\checkmark$   & \textbf{26.8}   & \textbf{112.5} & \textbf{87.8} & \textbf{26.0}      \\ 
\hline
\end{tabular}
}
\end{table}

\textbf{Video Multimodal Interpretation.}
As shown in~\cref{tab_3}, our approach outperforms existing video multimodal interpretation methods, achieving AP scores of $26.0$ and $28.7$ on the LV-VIS and OVIS datasets, respectively for class-level alignment.
This performance gain stems from the integration of the proposed GI-QFormer, which effectively captures complex cross-modal correlations between textual and visual features, along with the incorporation of the fine-grained alignment loss $L_{FA}$ and the multi-entity contrastive loss $L_{MC}$. These components jointly enforce consistent representations between corresponding caption–mask pairs, thereby enhancing the mask classification accuracy. 
In contrast, VideoGLaMM relies solely on text-level referring features for mask decoding. This limited modality interaction leads to suboptimal performance due to inherent cross-modal discrepancies and visual ambiguities between similar-looking objects, which hinder accurate pixel-level segmentation.
Notably, even when adopting Mask2Former as the segmentation backbone, our model still surpasses the OVFormer and OW-VISCap baselines, both of which also utilize Mask2Former, further demonstrating the robustness and effectiveness of our framework.
Moreover, our method implements the instance-level cross-modal alignment, with AP scores of $24.8$ and $27.5$ on the LV-VIS and OVIS datasets, respectively, which is approaching to the accuracy of the class-level results, demonstrating the effectiveness of the revised decoder.  
Since OW-VISCap \cite{choudhuri2024ow} has not released its code, so its results in \cref{tab_3} are cited from the references.

\subsection{Qualitative Results} 
\cref{fig:results} presents the quantitative results of our method in comparison with several advanced methods. As shown in \cref{fig:results} (a) and (b), OVFormer and SAM2 can only generate segmentation results. Furthermore, OVFormer can only associate its output masks with coarse-grained class labels, lacking fine-grained instance-level alignment.
In \cref{fig:results} (c), SMOTer generates target-specific descriptions for each object in the videos, but it has limited control over video segmentation, preventing it from tailoring the results to user needs. Moreover, SMOTer’s captions fail to align with specific targets in the videos, resulting in descriptions of the object’s state and actions without indicating its precise location.
In~\cref{fig:results}(d), VideoGLaMM produces overly simplistic captions and segmentation masks with poorly defined boundaries. When the initial captioning stage fails to correctly identify a target (e.g., the chainsaw or cabin in~\cref{fig:results}(d) left), the cascaded framework prevents the subsequent segmentation module from generating the corresponding mask. 
Moreover, VideoGLaMM relies solely on text-level referring features for mask decoding, and processes each target’s textual features independently within the segmentation module. This design not only amplifies inherent cross-modal discrepancies between textual and visual modalities but also aggravates ambiguities among visually similar objects. Consequently, the limited interaction across modalities and targets results in suboptimal performance, hindering accurate boundary delineation (e.g., adjacent ships) and leading to target loss over time (e.g., the disappearing log).
In contrast, as shown in~\cref{fig:results}(e), our method generates multiple coherent object masks along with a corresponding video caption based on a single user-provided box prompt. Furthermore, the integration of $L_{FA}$ and $L_{MC}$ ensures that noun entities in the generated captions are accurately aligned with their corresponding segmented objects, thereby ensuring tight cross-modal grounding. 
Overall, our approach delivers customized multimodal video interpretations guided by a user’s visual prompt, not only describing the actions associated with the specified target but also localizing semantically related objects in the scene.

\begin{table}[t!]
    \centering
    \caption{Performance comparison by using  different components of losses tested on the LV-VIS Val dataset.}
    \label{tab_5}
    \centering
\renewcommand\arraystretch{1.0}
\setlength\tabcolsep{2.5mm}
\resizebox{0.95\linewidth}{!}{
\begin{tabular}{cc|cc|c|c}
\hline 
$L_{FA}$    & $L_{MC}$    & SPICE & CIDEr & {${J}\&{F}$} & AP \\ \midrule
-            & -            & 23.8     & 106.9 & 84.5    & 22.1    \\
$\checkmark$            & - & 25.4     & 109.8 & 86.3    & 25.6    \\ 
- & $\checkmark$            & 25.1     & 110.4 & 86.9    & 25.8    \\
$\checkmark$ & $\checkmark$ & \textbf{26.8}   & \textbf{112.5} & \textbf{87.8}  & \textbf{26.0}    \\ 
\hline
\end{tabular}
}

\end{table}

\begin{table}[t!]
    \centering
    \caption{Performance comparison by using $\mathcal{L_{\text{MC}}}$ with different weights tested on the LV-VIS Val dataset.}
    \label{tab_6}
    \centering
\renewcommand\arraystretch{1.0}
\setlength\tabcolsep{2.5mm}
\resizebox{0.95\linewidth}{!}{
\begin{tabular}{l|ccccc}
\hline
$\lambda$ & 0 & 1 & 2 & 5 & 10 \\ \midrule
SPICE & 23.8 & 25.4 & \textbf{26.8} & 26.3 & 25.7 \\
CIDEr & 106.9 & 110.6 & \textbf{112.5} & 112.0 & 111.3 \\
{${J}\&{F}$} & 84.5 & 85.7 & \textbf{87.8} & 86.5 & 86.1 \\
 AP & 22.1 & 25.3 & \textbf{26.0} & 25.9 & 25.4 \\ \hline
\end{tabular}
}

\end{table}

\subsection{Ablation Studies}
\textbf{Analysis of PTGFormer.}
This experiment evaluates the PTGFormer incorporating an adaptive prompt adaptor to methodically filters out superfluous nodes and edges. 
To quantitatively assess the efficacy of this adaptor, we conduct a comprehensive analysis encompassing both spatial correlations (refer to Eq. \eqref{eq:2} and \eqref{eq:3}) and temporal correlations (refer to Eq. \eqref{eq:5}), as delineated in \cref{tab_4}. 
The experimental outcomes substantiate that the integration of spatial correlations engenders a substantial performance augmentation. By ascribing distinct weightings to graph nodes contingent upon their spatial affiliations with the prompt object, this methodology enables the derivation of more refined and semantically aligned node representations, congruent with user intent. 
Similarly, the incorporation of temporal correlations manifests an appreciable performance enhancement, facilitating the dynamic encoding of semantic nuances across sequential frames. 
The combined use of both correlations leads to substantial improvements in both segmentation and captioning tasks.

\textbf{Analysis of Alignment Losses.}
We conduct an ablation study to evaluate the efficacy of the proposed losses including $\mathcal{L}_{\text{FA}}$ and $\mathcal{L}_{\text{MC}}$, as illustrated in \cref{tab_5}. 
The use of  $\mathcal{L}_{\text{FA}}$ or $\mathcal{L}_{\text{MC}}$ both enhance the model's performance. 
Specifically, $\mathcal{L}_{\text{MC}}$ enforces the feature representation alignment between each mask and its corresponding caption word, thereby establishing a robust correspondence for robust decoding. 
Meanwhile, $\mathcal{L}_{\text{FA}}$ explicitly aligns the predicted masks with their corresponding instance words by modeling the mask-to-word distribution, corroborating the efficacy of the fine-grained alignment losses. \cref{tab_6} further provides additional evidence of the effectiveness of $\mathcal{L}_{\text{MC}}$ under different $\lambda$ in Eq. \eqref{eq:total_loss}. The optimal performance is achieved when $\lambda = 2$. Increasing $\lambda$ leads to a performance degradation as it diminishes the contributions of both the captioning and segmentation losses. Conversely, reducing $\lambda$ weakens the cross-modal alignment, resulting in a decline in overall performance.

\begin{figure*}[t!]
	\centering
\includegraphics[width=1\linewidth]{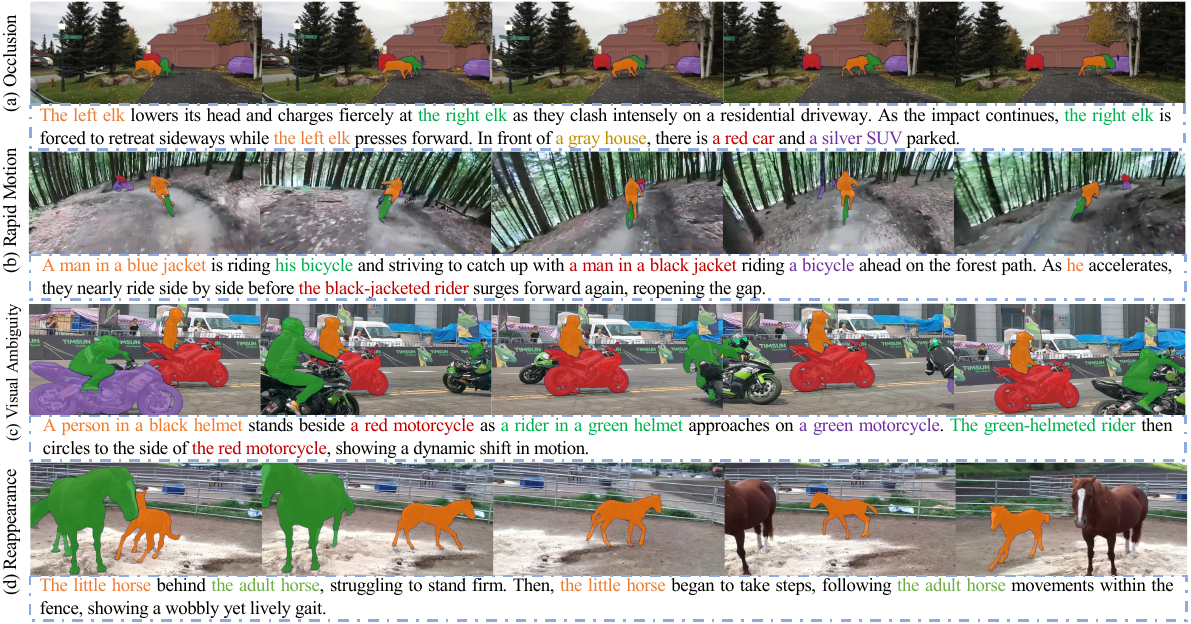}

	\caption{Visualization of several challenging and failure cases, including heavy occlusion, motion blur, visual ambiguity, and long-term target disappearance.}
	\label{fig:failure_cases}
\end{figure*}

\begin{table}[t!]
    \centering
    \caption{Performance comparison of three SG-FSCFormer variants, analyzing the impact of removing the prompt box, replacing the LLM, and using different mask decoders.}
    \label{ablation_variants}
    \centering
\renewcommand\arraystretch{1.0}
\setlength\tabcolsep{1.8mm}
\resizebox{1\linewidth}{!}{
\begin{tabular}{lcccc}
\hline 
 Method    & SPICE & CIDEr & {${J}\&{F}$} & AP \\ \midrule
Our (Vicuna-7B\&SAM2) & \textbf{26.8}   & \textbf{112.5} & \textbf{87.8}  & \textbf{26.0}    \\
\hline
\rowcolor[gray]{.9} 
\multicolumn{5}{c}{{\textit{Without Prompt Box}}}   \\ 
Our (w/o Prompt) & 25.4   & 110.8 & 86.5  & 25.3     \\
\hline
\rowcolor[gray]{.9} 
\multicolumn{5}{c}{{\textit{Using Different LLMs}}}   \\
Our (OPT-2.7B) & 25.9     & 111.2 & 87.4    & 25.7    \\
\hline
\rowcolor[gray]{.9} 
\multicolumn{5}{c}{{\textit{Using Different Mask Decoders}}}   \\
Our (Mask2Former) & 26.3   & 112.0 & 85.9  & 25.2    \\ 
\hline
\end{tabular}
}
\end{table}

\textbf{Effect of Prompt, LLM, and Mask Decoder.}
To evaluate the contribution of key components, we construct three variants of our model that respectively examine the impact of removing the prompt box, replacing the LLM backbone, and using different mask decoders.

Without Prompt Box. Since our task emphasizes controllable outputs, it is required to receive users' prompt and then capture desired results. This ensures that the output space aligns with user intent. 
As shown in \cref{ablation_variants} (“w/o Prompt”), the performance degrades when the prompt box is removed. This suggests that, in the absence of explicit spatial guidance, the model resorts to global semantic modeling, generating general descriptions and masks that may include irrelevant background objects or miss user-intended targets.
In practice, multiple prompt boxes can be employed to generate diverse outputs and enrich the output space; however, this lies beyond the scope of the current study.

Using Different LLMs.
We evaluated the impact of replacing the Vicuna-7B backbone with OPT-2.7B \cite{zhang2022opt}, and report the results in \cref{ablation_variants} (“Using Different LLMs”). 
When replacing Vicuna-7B with OPT-2.7B, we observe a performance drop across all metrics, indicating that stronger language modeling capabilities enhance semantic understanding in our framework.
Furthermore, as shown in \cref{tab_1}, Our model, whether equipped with Vicuna-7B or OPT-2.7B, consistently outperforms the MA-LMM baseline (which also uses Vicuna-7B) on the caption generation task. We attribute these gains to the synergistic effect between our scene-graph guided temporal semantic modeling and the integration of multimodal outputs.

Using Different Mask Decoders. 
To assess the impact of the SAM2 pretrained model in our framework, we replace it with Mask2Former, the same mask decoder used by the baseline methods.
As shown in \cref{tab_3} and \cref{ablation_variants}, our model equipped with Mask2Former still outperforms the baselines OVFormer and OW-VISCap, which also adopt Mask2Former. Although the performance decreases compared to that using SAM2, these results further demonstrate the effectiveness and generalizability of our framework.

\subsection{Challenging Cases Visualization}
We visualize several challenging and failure cases in \cref{fig:failure_cases} for detailed analysis, including occlusion, blurred motion, visual ambiguity, and long-term disappearance.
As shown in \cref{fig:failure_cases}(a) and (b), when the target undergoes occasional occlusion and motion blur, our method successfully produces accurate captions and corresponding masks.
In \cref{fig:failure_cases}(c), where the target exhibits highly similar visual features and temporarily disappears, our method erroneously segments a newly appearing rider upon the target’s return. 
We attribute this failure to interference caused by the high visual similarity between the original and newly appearing targets after the original object was lost.
In \cref{fig:failure_cases}(d), our method fails to segment the reappearing adult horse after a long-term disappearance. 
We believe this is due to the prolonged absence of the target, which overwrote the memory features maintained in the temporal buffer, thereby hindering effective re-identification and segmentation upon reappearance.
We plan to address these limitations in future work.

\subsection{Inference Speed and Model Parameters}
\label{sec:Complexity_Analysis}
In our SG-FSCFormer, the trainable modules include PTGFormer, and the MLDecoder, with their respective parameters and FLOPs detailed in \cref{tab:param}. 
Based on SAM2 and Vicuna-7B, our approach only adds PTGFormer (10.2M) and MLDecoder (85.7M), and achieves 4.96 FPS tested on an A6000 GPU. 
Furthermore, we compare the inference speed of our method with OVFormer, SAM2, MA-LMM, SMOTer, and  VideoGLaMM, which achieve 2.98, 10.61, 7.35, 4.82, and 5.13 FPS, respectively, as shown in \cref{tab:speed}.
While the inference speed of our model is lower than unimodal generation approaches such as SAM2 for segmentation and MA-LMM for captioning, it outperforms SMOTer, which jointly generates bounding boxes and captions, and achieves inference speed comparable to VideoGLaMM.
Overall, our approach introduces relatively few learnable parameters and delivers efficient inference, achieving a favorable balance between model complexity and effectiveness.

\begin{table}[t!]
    \caption{Parameters and FLOPs of the trainable modules in SG-FSCFormer.}
    \label{tab:param}
    \renewcommand\arraystretch{1.0}
\centering
\setlength\tabcolsep{6.5mm}
\resizebox{1.0\linewidth}{!}{
\begin{tabular}{lcc}
\toprule
Module & PTGFormer & MLDecoder\\
\midrule    
Params & 10.2M & 85.7M \\
FLOPs  & 1.8G & 3.1G \\
\bottomrule
\end{tabular}
}
\end{table}

\begin{table}[t!]
    \centering
    \caption{Inference speed (FPS) comparison.}
    \label{tab:speed}
    \centering
\renewcommand\arraystretch{1.0}
\setlength\tabcolsep{1.0mm}
\resizebox{1.0\linewidth}{!}{
\begin{tabular}{c|ccccc|c}
\hline
Method   & OVFormer & SAM2  & MA-LMM & SMOTer & VideoGLaMM  & Ours \\ \midrule
FPS   & 2.98  & 10.61 & 7.35 & 4.82 & 5.13 & 4.96  \\
\hline
\end{tabular}}

\end{table}

\subsection{Limitations}
\label{sec:limitations}
Current video segmentation and captioning methods generate unimodal outputs, limiting the user's ability to access rich multimodal data. Additionally, the integration of video segmentation and captioning often results in oversimplification or omission of content that may be of interest to the user. In contrast, our SegCaptioning task leverages user prompts as guidance and incorporates temporal scene graph modeling, which effectively captures the contextual content of interest and collaboratively generates both video segmentation and contextual descriptions.
However, our method relies on manually annotated alignment of masks and captions in two existing video datasets. Scaling this method to larger datasets (e.g., at an Internet scale) remains challenging due to the scarcity of annotated data and high labeling costs. 
Future work will explore human-supervised semi-automated annotation pipelines and weakly supervised pretraining to overcome this limitation.
Additionally, as illustrated in \cref{fig:failure_cases}(c) and (d), our method may struggle with targets that exhibit high visual similarity or remain absent from the scene for extended periods.
We plan to mitigate these issues by optimizing the memory features maintained in the temporal buffer, enabling more robust long-range modeling and improved target discrimination.

\section{Conclusion}  \label{conclusion}
This paper introduces a novel research task, ``Controllable Video Segmentation and Captioning'', establishing it as the inaugural controllable video multimodal interpretation challenge and offering valuable insights to inform future research endeavors. To tackle this task, we propose an innovative framework, ``Scene Graph-guided Fine-grained SegCaptioning Transformer'', which incorporates a PTGFormer to effectively translate a simple user prompt into prompt-specific graphs, meticulously aligned with user intent. Furthermore, the proposed Mask-linguistic Decoder explicitly enforces alignment between each mask and its corresponding caption tokens, implementing fine-grained alignment to generate precise multimodal outputs that enhance user comprehension of video content. Extensive empirical evaluations on two benchmark datasets substantiate the efficacy of our method, demonstrating its capacity to accurately capture user intent and generate robust multimodal outputs tailored to user needs.

{
\bibliographystyle{IEEEtran}
\bibliography{main}
}

\end{document}